# 3D-Mesh denoising using an improved vertex based anisotropic diffusion

Mohammed El Hassouni
DESTEC
FLSHR, University of Mohammed V-Agdal-
Rabat, Morocco
Mohamed.Elhassouni@gmail.com

Driss Aboutajdine
LRIT, UA CNRST
FSR, University of Mohammed V-Agdal-
Rabat, Morocco
aboutaj@fsr.ac.ma

*Abstract*—This paper deals with an improvement of vertex based nonlinear diffusion for mesh denoising. This method directly filters the position of the vertices using Laplace, reduced centered Gaussian and Rayleigh probability density functions as diffusivities. The use of these PDFs improves the performance of a vertex-based diffusion method which are adapted to the underlying mesh structure. We also compare the proposed method to other mesh denoising methods such as Laplacian flow, mean, median, min and the adaptive MMSE filtering. To evaluate these methods of filtering, we use two error metrics. The first is based on the vertices and the second is based on the normals. Experimental results demonstrate the effectiveness of our proposed method in comparison with the existing methods.

*Keywords*- Mesh denoising, diffusion, vertex.

## I. INTRODUCTION

The current graphic data processing tools allow the design and the visualization of realistic and precise 3D models. These 3D models are digital representations of either the real world or an imaginary world. The techniques of acquisition or design of the 3D models (modellers, scanners, sensors) generally produce sets of very dense data containing both geometrical and appearance attributes. The geometrical attributes describe the shape and dimensions of the object and include the data relating to a unit of points on the surface of the modelled object. The attributes of appearance contain information which describes the appearance of the object such as colours and textures.

These 3D models can be applied in various fields such as the medical imaging, the video games, the cultural heritage... etc [1]. These 3D data are generally represented by polygonal meshes defined by a unit of vertex and faces. The most meshes used for the representation of objects in 3D space are the triangular surface meshes.

The presence of noise in surfaces of 3D objects is a problem that should not be ignored. The noise affecting these surfaces can be topological, therefore it would be created by algorithms used to extract the meshes starting from groups of vertices; or geometrical, and in this case it would be due to the errors of measurements and sampling of the data in the various treatments [2].

To eliminate this noise, a first study was made by Taubin [3] by applying signal processing methods to surfaces of 3D objects. This study has encouraged many researchers to develop extensions of image processing methods in order to apply them to 3D objects. Among these methods, there are those based on Wiener filter [4], Laplacian flow [5] which adjusts simultaneously the place of each vertex of mesh on the geometrical center of its neighboring vertex, median filter [5], and Alpha-Trimming filter [6] which is similar to the nonlinear diffusion of the normals with an automatic choice of threshold.
The only difference is that instead of using the nonlinear average, it uses the linear average and the non iterative method based on robust statistics and local predictive factors of first order of the surface to preserve the geometric structure of the data [7].

There are other approaches for denoising 3D objects such as adaptive filtering MMSE [8]. This filter depends on the form [9] which can be considered in a special case as an average filter [5], a min filter [9], or a filter arranged between the two. Other approaches are based on bilateral filtering by identification of the characteristics [10], the non local average [11] and adaptive filtering by a transform in volumetric distance for the conservation of the characteristics [12].

Recently, a new method of diffusion based on the vertices [13] was proposed by Zhang and Ben Hamza. It consists in solving a nonlinear discrete partial differential equation by entirely preserving the geometrical structure of the data.

In this article, we propose an improvement of the vertex based diffusion proposed by Zhang and Ben Hamza. The only difference is to use of different diffusivities such as the functions of Laplace, reduced centred Gaussian and Rayleigh instead of the function of Cauchy. To estimate these various methods of filtering, two error metric $L^2$ [13] are used.

This article is organized as follows: Section 2 presents the problem formulation. In Section 3, we review some 3D mesh denoising techniques; Section 4 presents the proposed approaches; Section 5 presents the used error metrics. In Section 6, we provide experimental results to demonstrate a much improved performance of the proposed methods in 3D mesh smoothing. Section 7 deals with some concluding remarks.

## II. PROBLEM FORMULATION

3D objects are usually represented as polygonal or triangle meshes. A triangle mesh is a triple $M = (P, \varepsilon, T)$ where $P = \{P_1, ......, P_n\}$ is the set of vertices, $\varepsilon = \{e_{ij}\}$ is the set of edges and $T = \{T_1, ......, T_n\}$ is the set of triangles. Each edge connects a pair of vertices $(P_i, P_j)$. The neighbouring of a vertex is the set $P^* = \{P_j \in P: P_i \sim P_j\}$. The degree $d_i$ of a vertex $P_i$ is the number of the neighbours $P_j$. $N(P_i)$ is the set of the neighbouring vertices of $P_i$. $N(T_i)$ is the set of the neighbouring triangles of $T_i$.

We denote by $A(T_i)$ and $n(T_i)$ the area and the unit normal of $T_i$, respectively. The normal $n$ at a vertex $P_i$ is obtained by averaging the normals of its neighbouring triangles and is given by

$$n_i = \frac{1}{d_{i5}} \sum_{T_j \in T(P_i^*)} n(T_j) \quad (1)$$

The mean edge length $\bar{l}$ of the mesh is given by

$$\bar{l} = \frac{1}{|\varepsilon|} \sum_{e_{ij} \in \varepsilon} \|e_{ij}\| \quad (2)$$

During acquisition of a 3D model, the measurements are perturbed by an additive noise:

$$P = P' + \eta \quad (3)$$

Where the vertex $P$ includes the original vertex $P'$ and the random noise process $\eta$. This noise is generally considered as a Gaussian additive noise.

For that, several methods of filtering of the meshes were proposed to filter and decrease the noise contaminating the 3D models.

## III. RELATED WORK

In this section, we present the methods based on the normals such as the mean, the median, the min and the adaptive MMSE filters and the methods based on the vertices such as the laplacien flow and the vertex-based diffusion using the functions of Cauchy, Laplace, Gaussian and Rayleigh.

### A. Normal-based methods

Consider an oriented triangle mesh. Let $T$ and $U_i$ be a mesh triangles, $n(T)$ and $n(U_i)$ be the unit normal of $T$ and $U_i$ respectively, $A(T)$ be the area of $T$, and $C(T)$ be the centroid of $T$. Denote by $N(T)$ the set of all mesh triangles that have a common edge or vertex with $T$ (see Fig. 1).

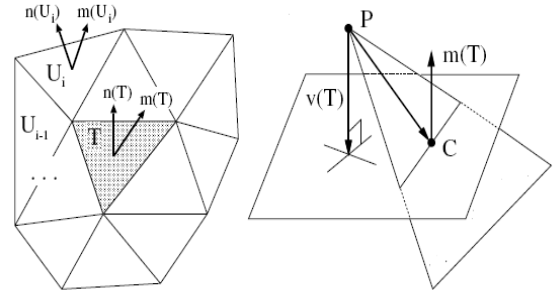

Fig. 1 Left: Triangular mesh.
Right: updating mesh vertex position.

*1) Mean Filter:* The mesh mean filtering scheme includes three steps [5]:

**Step 1.** For each mesh triangle $T$, compute the averaged normal $m(T)$ :

$$m(T) = \frac{1}{\sum A(U_i)} \sum_{U_i \in N(T)} A(U_i) n(U_i) \quad (4)$$

**Step 2.** Normalize the averaged normal $m(T)$ :

$$m(T) \leftarrow \frac{m(T)}{\|m(T)\|} \quad (5)$$

**Step 3.** Update each vertex in the mesh:

$$P_{new} \leftarrow P_{old} + \frac{1}{\sum A(T)} \sum A(T) v(T) \quad (6)$$

With

$$v(T) = \left|\overrightarrow{PC}..m(T)\right| m(T) \quad (7)$$

$v(T)$ is the projection of the vector $\overrightarrow{PC}..$ onto the direction of $m(T)$, as shown by the right image of Fig. 1.

*2) Min filter :* The process of min filtering differs from the average filtering only at step1. Instead of making the average of the normals, we determine the narrowest normal, $\eta_i$, for each face, by using the following steps [9]:

- Compute of angle $\Phi$ between $n(T)$ and $n(U_i)$.

- Research of the minimal angle: If $\Phi$ is the minimal angle in $N(T)$ then $n(T)$ is replaced by $n(U_i)$.

*3) Angle Median Filter:* This method is similar to min filtering; the only difference is that instead of seeking the narrowest normal we determine the median normal by applying the angle median filter [5]:

$$\theta_i = \angle(n(T), n(U_i)) \quad (8)$$

If $\theta_i$ is the median angle in $N(T)$ then $n(T)$ is replaced by $n(U_i)$.

*4) Adaptive MMSE Filter:* This filter differs from the average filter only at step1. The new normal m(T) for each triangle T is calculated by [8]:

$$m_j(T) = \begin{cases} M_{lj}(T) & \sigma_n^2 > \sigma_{lj}^2 \text{ ou } \sigma_{lj}^2 = 0 \\ \left(1 - \frac{\sigma_n^2}{\sigma_{lj}^2}\right) n_j(T) + \frac{\sigma_n^2}{\sigma_{lj}^2} M_{lj}(T) & \sigma_n^2 \leq \sigma_{lj}^2 \text{ et } \sigma_{lj}^2 \neq 0 \end{cases} \quad (9)$$

$$M_{lj}(T) = \frac{\sum_{i=0}^{N-1} A(U_i) n_j(U_i)}{\sum_{i=0}^{N-1} A(U_i)} \quad (10)$$

$\sigma_n^2$ is the variance of additive noise and $\sigma_{lj}^2$ is the variance of neighbouring mesh normals which is changed according to elements of normal vector. Thus, $\sigma_{lj}^2$ is calculated as follows:

$$\sigma_{lj}^2 = \frac{\sum_{i=0}^{N-1} A(U_i) n_j^2(U_i)}{\sum_{i=0}^{N-1} A(U_i)} - M_{lj}^2(T) \quad (11)$$

B. *Vertex-based methods*
  *1) Laplacian Flow*

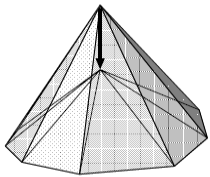

Fig 2. Updating vertex position by umbrella operator.

Considering the following expression which allows the update of the mesh vertices [5]

$$P_{new} \leftarrow P_{old} + \lambda D(P_{old}) \quad (12)$$

Where $D(P)$ is a displacement vector, and $\lambda$ is a step-size parameter.

The Laplacian smoothing flow is obtained if the displacement vector $D(P)$ is defined by the so-called umbrella operator [14] (see Fig. 2) :

$$U(P_i) = \frac{1}{n} \sum_{j \in N(P)} P_j - P_i \quad (13)$$

$N(P)$ is the 1-ring of mesh vertices neighbouring on $P_i$.

*2) Vertex-Based Diffusion using the Function of Cauchy:* This method [13] consists in updating the mesh vertices by solving a nonlinear discrete partial differential equation using the function of Cauchy.

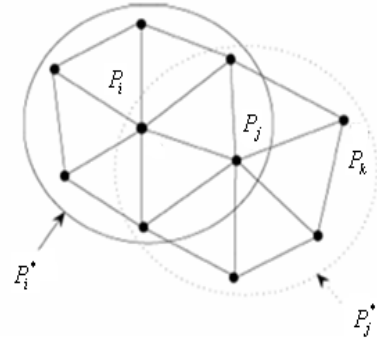

Fig 3. Illustration of two neighbouring rings.

The update of the vertices of mesh (see Fig. 3) is given by

$$P_i \leftarrow P_i + \sum_{P_j \in P_i^*} \frac{1}{\sqrt{d_i}} \left( \frac{P_j}{\sqrt{d_j}} - \frac{P_i}{\sqrt{d_i}} \right) \left( g(|\nabla P_i|) + g(|\nabla P_j|) \right) \quad (14)$$

Where $g$ is Cauchy weight function given by

$$g(x) = \frac{1}{1 + \frac{x^2}{c^2}} \quad (15)$$

and $c$ is a constant tuning parameter that needs to be estimated.

The gradient magnitudes are given by

$$|\nabla P_i| = \left(\sum_{P_j \in P_i^*} \left\|\frac{P_i}{\sqrt{d_i}} - \frac{P_j}{\sqrt{d_j}}\right\|^2\right)^{1/2} \quad (16)$$

And

$$|\nabla P_j| = \left(\sum_{P_k \in P_j^*} \left\|\frac{P_j}{\sqrt{d_j}} - \frac{P_k}{\sqrt{d_k}}\right\|^2\right)^{1/2} \quad (17)$$

Note that the update rule of the proposed method requires the use of two neighbouring rings as depicted in Fig. 3.

## IV. PROPOSED METHOD

The method of vertex-based diffusion [13] was proposed by Zhang and Ben Hamza and which consists in solving a nonlinear discrete partial differential equation using the function of Cauchy.

In this section, we propose an improvement of the vertex-based diffusion proposed by Zhang and Ben Hamza. The only difference is the use of other diffusivity functions instead of Cauchy function. These functions are presented as follows:

- Reduced Centered Gaussian function :

$$g(x) = \sqrt{\frac{1}{2 \times pi}} \times \exp\frac{-\left(\frac{x}{c}\right)^2}{2} \quad (18)$$

- Laplace function :

$$g(x) = \frac{\exp\left(-abs\left(\frac{x}{c}\right)\right)}{2} \quad (19)$$

- Rayleigh function :

$$g(x) = \exp\left(-\frac{\left(\frac{x}{c}\right)^2}{2}\right) \times \left(\frac{x}{c}\right) \quad (20)$$

$c$ is a constant tuning parameter that needs to be estimated for each distribution.

## V. $L^2$ ERROR METRIC

To quantify the better performance of the proposed approaches in comparison with the method based on the vertices using the function of Cauchy and the other methods, we computed the vertex-position and the face-normal error metrics $L^2$ [13].

Consider an original model $M'$ and the model after adding noise or applying several iterations smoothing $M$. $P$ is a vertex of $M$. Let set *dist (P, M')* equal to the distance between $P$ and a triangle of the ideal mesh $M'$ closest to $P$. Our $L^2$ vertex-position error metric is given by

$$\varepsilon_v = \frac{1}{3A(M)} \sum_{P \in M} A(P) dist(P, M')^2 \quad (21)$$

Where $A(P)$ is the summation of areas of all triangles incident on $P$ and $A(M)$ is the total area of $M$.

The face-normal error metric is defined by

$$\varepsilon_f = \frac{1}{A(M)} \sum_{T \in M} A(T) \|n(T') - n(T)\|^2 \quad (22)$$

Here $T$ and $T'$ are triangles of the meshes $M$ and $M'$ respectively; $n(T)$ and $n(T')$ are the unit normals of $T$ and $T'$ respectively and $A(T)$ is the total area of $T$.

## VI. EXPERIMENTAL RESULTS

This section presents simulation results where the normal based methods, the vertex-based methods and the proposed method are applied to noisy 3D models obtained by adding Gaussian noise as shown in Figs 6 and 8.
The standard deviation of Gaussian noise is given by

$$\sigma = noise \times \bar{l} \quad (23)$$

Where $\bar{l}$ is the mean edge length of the mesh.

We also test the performance of the proposed methods on original noisy laser-scanned 3D models shown in Figs 4 and 10.
The method of vertex-based diffusion using the proposed diffusivity functions of Laplace, reduced centred Gaussian and Rayleigh are a little bit more accurate than the method of vertex-based diffusion using the function of Cauchy. Some features are better preserved with the approaches of vertex based diffusion using these functions (see Figs 4 and 10).
By comparing the four distinct methods (see Figs 5 and 11), we notice that the proposed method gives the smallest error metrics comparing to method of vertex-based diffusion using the function of Cauchy.
The experimental results show clearly that vertex-based methods outperform the normal-based methods in term of visual quality. These results are illustrated by Fig 6.
In Fig 7, the values of the two error metrics show clearly that the vertex-based diffusion using the functions of Laplace, reduced centred Gaussian and Rayleigh give the best results and they are more effective than the methods based on the normals. Fig 7 also shows that the approaches based on the vertices such as Laplacien flow and the vertex-based diffusion using the functions of Cauchy, Laplace, reduced centred Gaussian and Rayleigh give results whose variation is remarkably small.

In all the experiments, we observe that the vertex-based diffusion using different laws is simple and easy to implement, and require only some iterations to remove the noise. The increase in the number of iteration involves a problem of over smoothing (see Fig 8). In Fig 9, we see that the method of vertex-based diffusion using the function of Cauchy leads more quickly to an over smoothing than the methods of vertex-based diffusion using the functions of Laplace, reduced centered Gaussian and Rayleigh.

## VII. CONCLUSION

In this paper, we introduced a vertex-based anisotropic diffusion for 3D mesh denoising by solving a nonlinear discrete partial differential equation using the diffusivity functions of Laplace, reduced centered Gaussian and Rayleigh. These method is efficient for 3D mesh denoising strategy to fully preserve the geometric structure of the 3D mesh data. The experimental results clearly show a slight improvement of the performance of the proposed approaches using the functions of Laplace, reduced centered Gaussian and Rayleigh in comparison with the methods of the laplacien flow and the vertex-based diffusion using the function of Cauchy. The Experiments also demonstrate that our method is more efficient than the methods based on the normals to mesh smoothing.

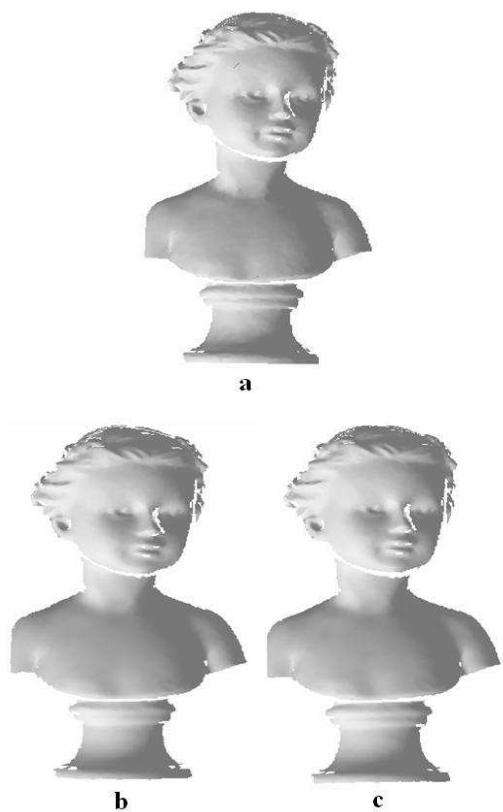

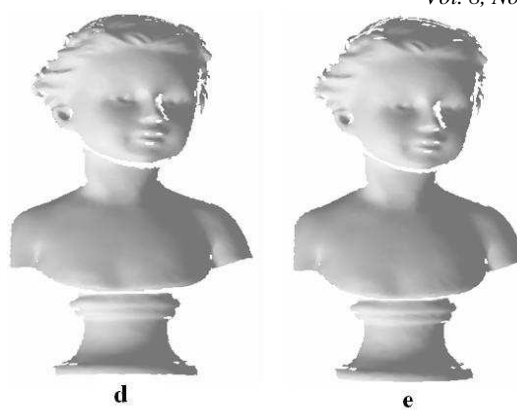

Fig 4. (a) Statue model digitized by a Roland LPX-250 laser range scanner (23344 vertices and 45113 faces); smoothing model by method based on the vertices using the functions of (b) Cauchy (c = 15.3849), (c) Laplace (c = 37.3849), (d) Gaussian (c = 37.3849) and (e) Rayleigh (c = 37.3849). The number of iteration times is 7 for each case.

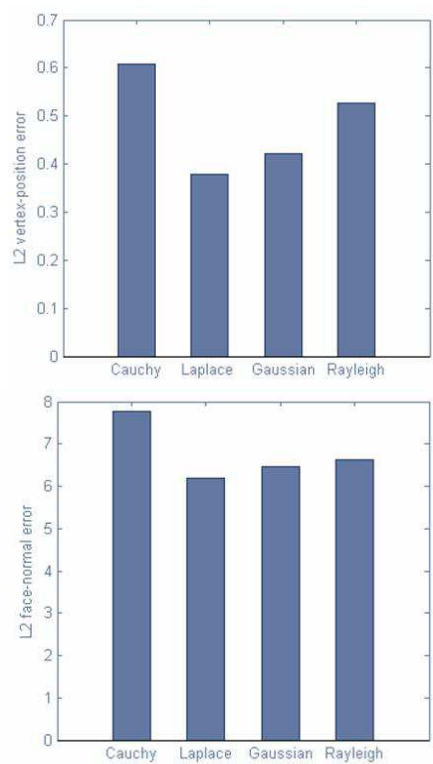

Fig 5. Top: $L^2$ vertex-position error metric of 3D model in Fig 4 Bottom: $L^2$ face-normal error metric of 3D model in Fig 4

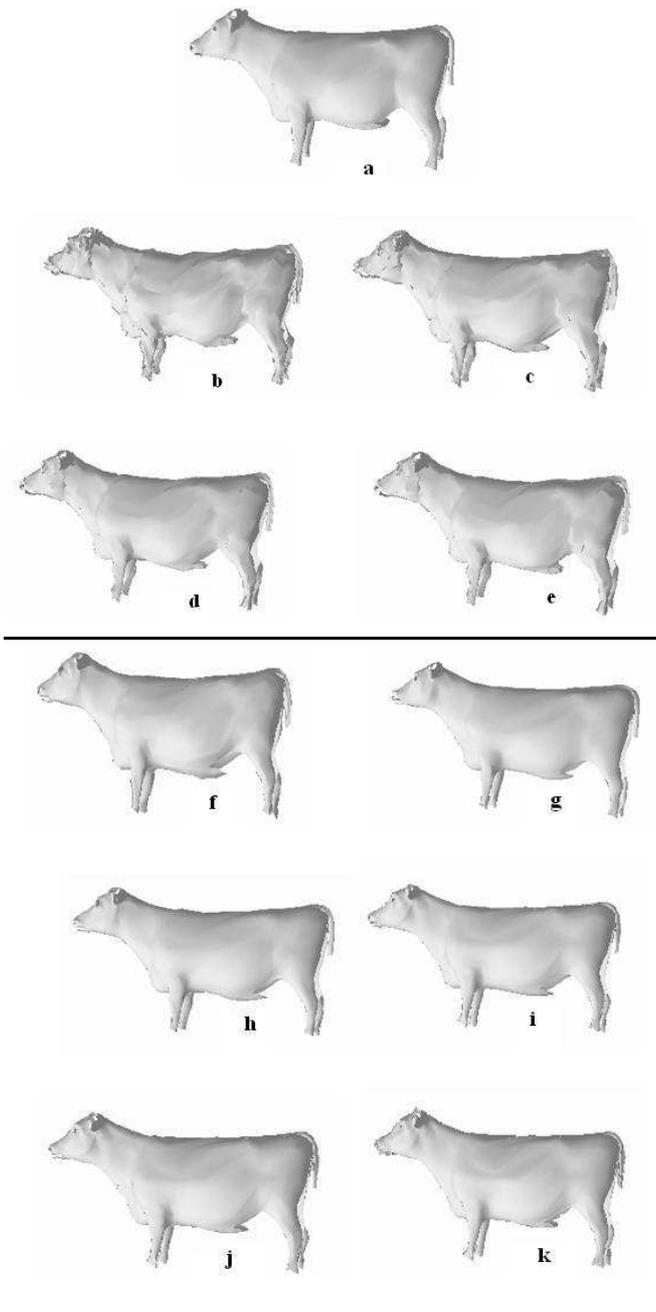

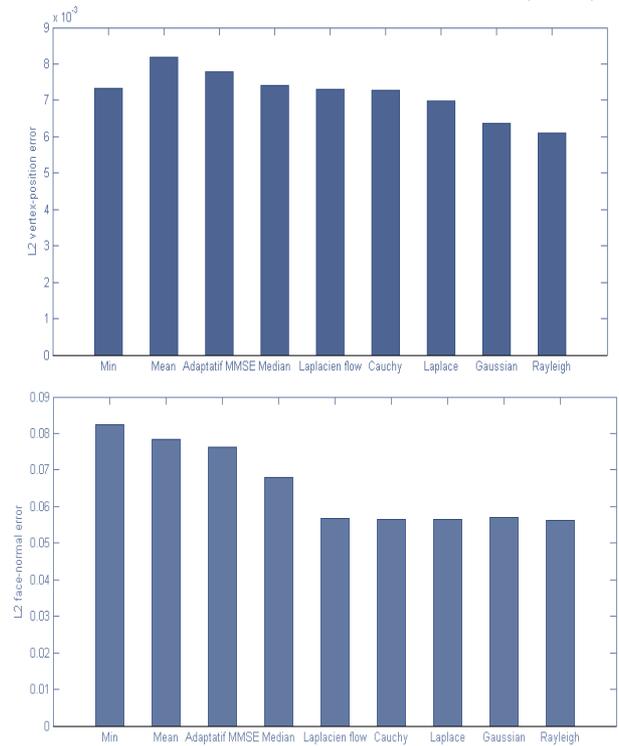

Fig7. Left: $L^2$ vertex-position error metric of 3D model in Fig 6. Right: $L^2$ face-normal error metric of 3D model in Fig 6.

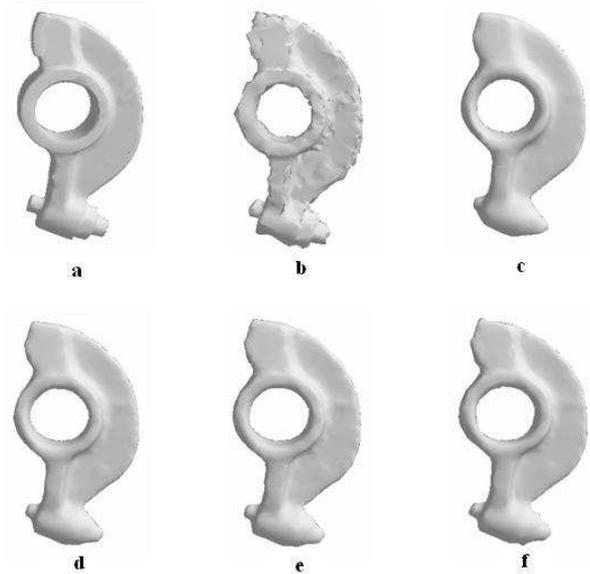

Fig 6. (a) Original model(4349 vertices and 2260 faces); (b) Adding Gaussian noise ($\varepsilon_v = 0.0090$, $\varepsilon_f = 0.0994$ and $\sigma = 0.8\ \bar{l}$ ); (c) Min filter (7 iterations); (d) Mean filter (3 iterations); (e) Adaptatif MMSE filter (3 iterations); (f) Median filter (4 iterations); (g) Laplacien flow (2 iterations and λ=0.45); smoothing model by method based on the vertices using the functions of (h) Cauchy (3 iterations and c = 2.3849), (i) Laplace (6 iterations and c = 8.3849), (j) Gaussian (6 iterations and c= 8.3849) and (k) Rayleigh (6 iterations and c = 0.3).

Fig 8. (a) Original model (2108 vertices and 4216 faces); (b) Adding Gaussian noise ($\sigma = 0.7\ \bar{l}$ ); smoothing model by method based on the vertices using the functions of (c) Cauchy (c = 2.3849), (d) Laplace (c = 15.3849), (e) Gaussian (c = 15.3849) and (f) Rayleigh (c = 0.03849). The number of iteration times is 10 for each case.

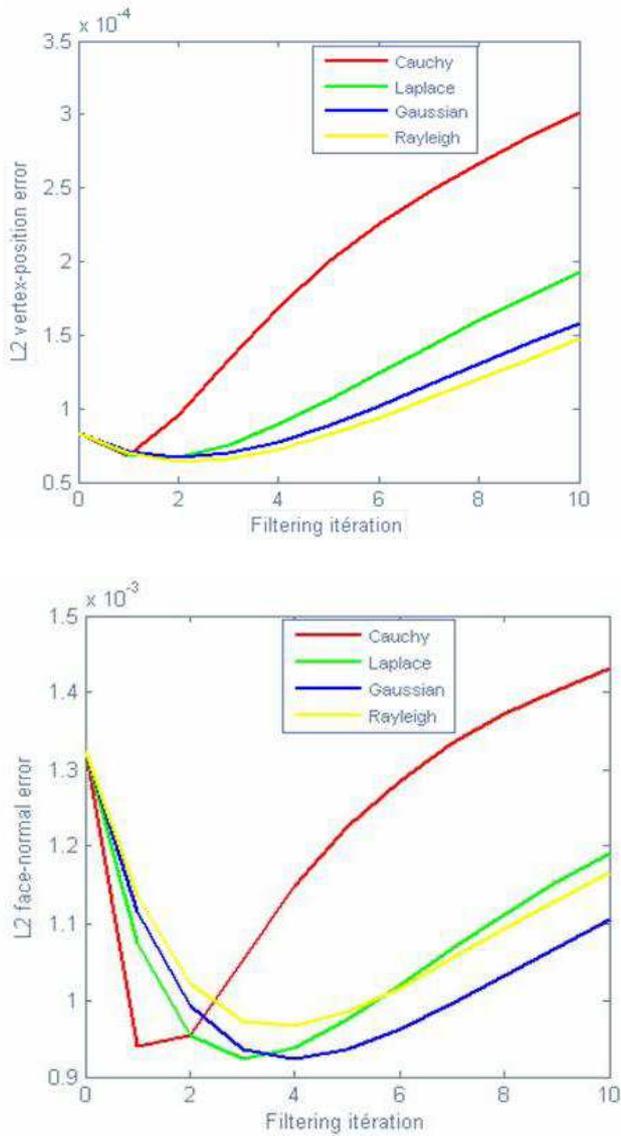

Fig 9. Left: $L^2$ vertex-position error metric of 3D model in Fig 8. Right: $L^2$ face-normal error metric of 3D model in Fig8.

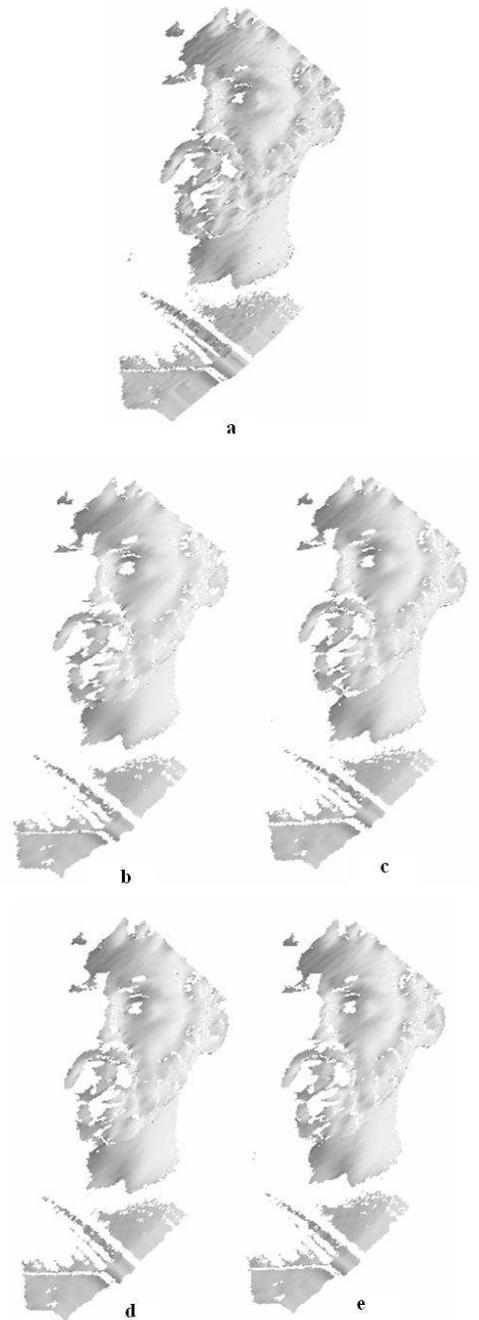

Fig 10. (a) Statue model digitized by impact 3D scanner (59666 vertices and 109525 faces); smoothing model by method based on the vertices using the functions of (b) Cauchy (c = 15.3849), (c) Laplace (c = 37.3849), (d) reduced centered Gaussian (c = 37.3849) and (e) Rayleigh (c = 37.3849).The number of iteration times is 11 for each case.

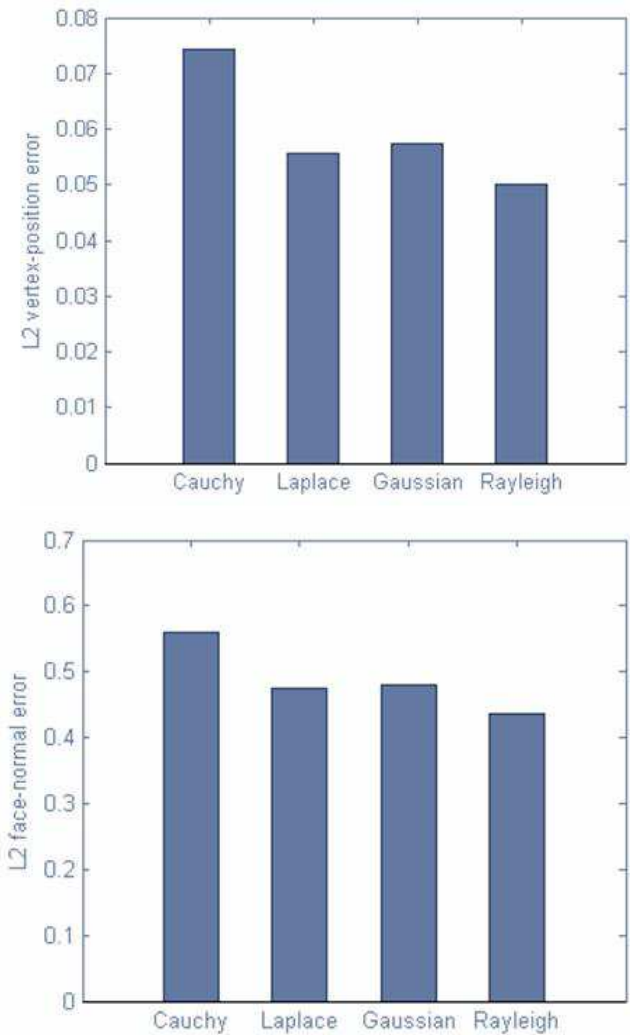

Fig 11. High: $L^2$ vertex-position error metric of Fig 10. Low: $L^2$ face-normal error metric of Fig 10.